\documentclass[letterpaper, 10 pt, conference]{ieeeconf}  %

\IEEEoverridecommandlockouts                              %

\title{\LARGE \bf
Imitation Learning with Limited Actions via Diffusion Planners and 
Deep Koopman Controllers
}
\author{Jianxin Bi$^{1}$, Kelvin Lim$^{1}$, Kaiqi Chen$^{1}$, Yifei Huang$^{1}$, and Harold Soh$^{1,2}$%
\thanks{$^{1}$Department of Computer Science, School of Computing, National University of Singapore, Singapore.}%
\thanks{$^{2}$ Smart Systems Institute, NUS.}%
\thanks{Contact Authors: {\tt\scriptsize  \{jianxin.bi, harold\}@comp.nus.edu.sg}}
}

\usepackage{amsmath}
\usepackage{amssymb}
\usepackage{microtype}
\usepackage{graphicx}
\usepackage{cite}
\usepackage{hyperref}
\usepackage{balance}
\usepackage[T1]{fontenc}

\usepackage{booktabs}  % For better table lines
\usepackage{multirow}  % For multi-row cells
\usepackage{siunitx} % for formatting numbers
\graphicspath{ {./figure/} }
\usepackage{xcolor}

\newcommand{\para}[1]{\vspace{0.1cm}\noindent\textbf{#1}}

\newcommand{\method}{KOAP}
\usepackage{hyperref}

\begin{document}

\maketitle
\thispagestyle{empty}
\pagestyle{empty}

\begin{abstract}

Recent advances in diffusion-based robot policies have demonstrated significant potential in imitating multi-modal behaviors. However, these approaches typically require large quantities of demonstration data paired with corresponding robot action labels, creating a substantial data collection burden. In this work, we propose a plan-then-control framework aimed at improving the action-data efficiency of inverse dynamics controllers by leveraging observational demonstration data. Specifically, we adopt a Deep Koopman Operator framework to model the dynamical system and utilize observation-only trajectories to learn a latent action representation. This latent representation can then be effectively mapped to real high-dimensional continuous actions using a linear action decoder, requiring minimal action-labeled data. Through experiments on simulated robot manipulation tasks and a real robot experiment with multi-modal expert demonstrations, we demonstrate that our approach significantly enhances action-data efficiency and achieves high task success rates with limited action data.

\end{abstract}

\section{Introduction}

Learning effective policies from demonstrations is a fundamental challenge in robotics and machine learning. Traditional imitation learning often relies on complete state-action trajectories, but obtaining action labels can be costly or impractical in many real-world scenarios. This limitation has spurred interest in \emph{imitation from observation}, where the goal is to learn policies primarily from state observations without corresponding action data. Successfully addressing this challenge would enable scalable learning systems capable of leveraging abundant unlabeled observational data to perform complex tasks.

\begin{figure}
    \includegraphics[scale=0.17]{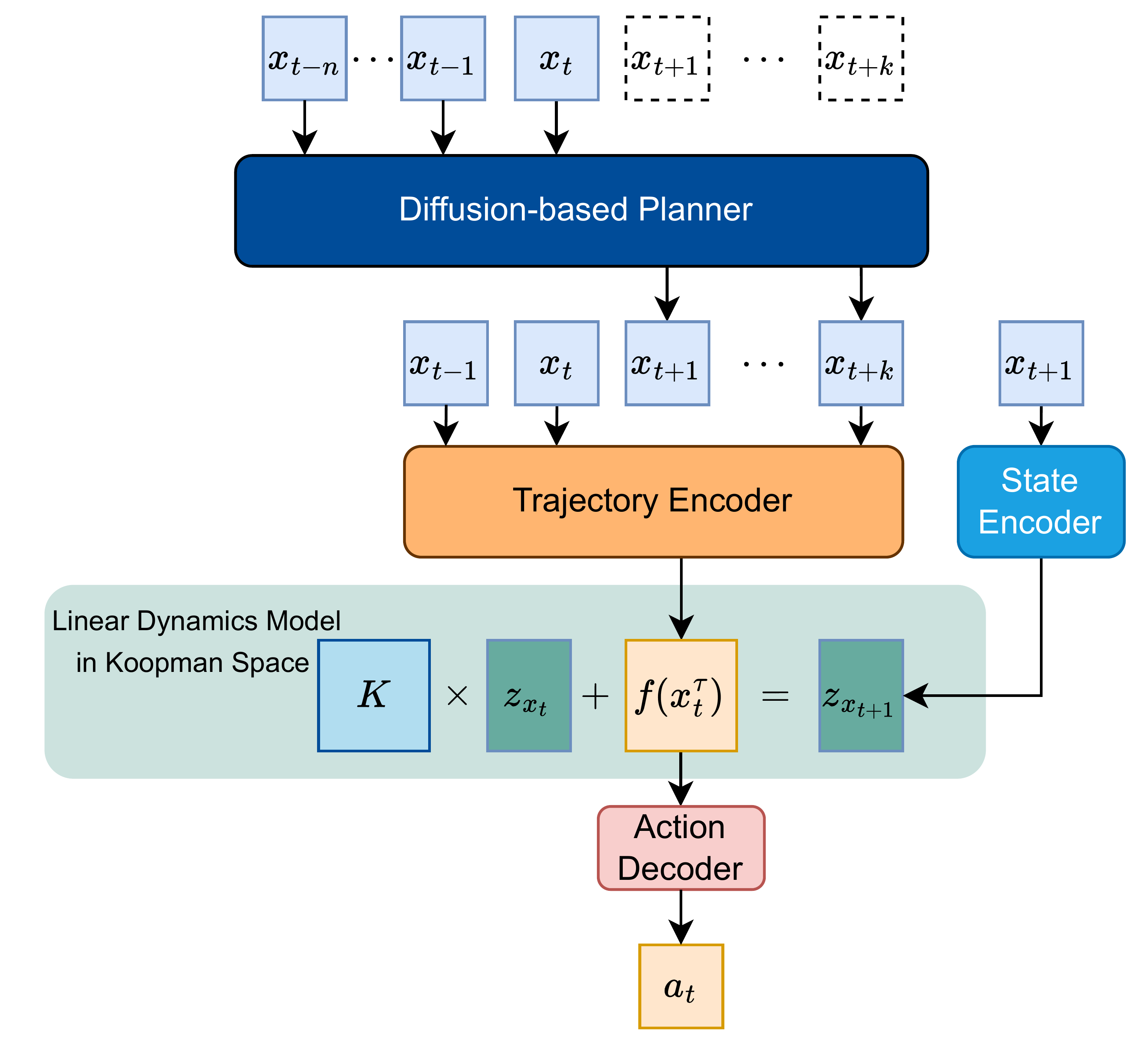}
    \caption{We adopt a plan-then-control scheme: a diffusion-based planner generates future states based on current and past states, with an inverse dynamics controller that generate action sequence to follow target trajectory. We propose \method{}, a method for leveraging action-free trajectories to improve controller learning with limited action data. \method{} exploits Deep Koopman Operators to lift the nonlinear target system into a linear latent space, which regularizes latent action learning. Real actions can be decoded through learning a simple (linear) action decoder by using action data.}
    \label{fig:wide-figure}
\end{figure}

In this work, we adopt a plan-then-control scheme with Decision Diffusers~\cite{ajay2023is} (Fig. \ref{fig:wide-figure}) and propose \emph{Koopman Operator with Action Proxy} (\method{}), a simple yet effective approach for learning an inverse dynamics controller with limited action data. Inspired by Deep Koopman Operators~\cite{Lusch2017DeepLF}, the key insight underlying \method{} is to learn useful \emph{latent actions} by regularizing the latent space with linear dynamics. By enforcing linear forward dynamics, the learned latent actions capture meaningful differences between lifted states, facilitating accurate action predictions even with limited action supervision. Applying a simple (linear) action decoder further minimizes overfitting risks when action-labeled data is scarce.

Our approach differs from prior works such as ILPO~\cite{edwards2019imitating} and LAPO~\cite{schmidt2024learning}, which focus primarily on discrete action policies and involve more complex modeling of latent actions and dynamics, often requiring information bottlenecks to retain essential information. In contrast, \method{} lifts the nonlinear system into a linear space of continuous latent states and latent actions, allowing us to naturally handle continuous action policies. By integrating decision diffusers for planning and focusing on learning an effective inverse model, we directly map latent actions to real actions without the need for complex architectures or quantization bottlenecks. To our knowledge, \method{} is the first work to combine decision diffusers with Koopman-inspired inverse model learning, providing a simple and scalable solution for imitation learning with limited action data.

We validate \method{} on a range of simulated robot manipulation tasks from the D3IL benchmark~\cite{jia2024towards}. Our experiments demonstrate that \method{} significantly outperforms baseline methods, particularly when action-labeled data is limited. We observe that \method{} effectively learns informative latent actions, resulting in strong policy performance. Performance improves with additional observational data, and \method{} can even learn useful latent representations without any action labels. A case-study for scooping on a real-robot system shows our approach outperforms existing methods.

By addressing the challenge of learning from observations with limited action data, \method{} advances the field of imitation from observations. Our key contributions are:
\begin{itemize}
    \item We introduce \method{}, a simple, scalable method that combines decision diffusers for planning with effective inverse model learning, leveraging Koopman Operator theory to regularize latent actions through linear latent dynamics and linear action decoding.
    \item We demonstrate that \method{} effectively learns informative latent actions, leading to strong policy performance even with limited action-labeled data.
    \item We show that \method{} outperforms existing methods and variants on complex manipulation tasks.
\end{itemize}
This work enables scalable and efficient policy learning from observations, reducing dependence on costly action labeling and paving the way for more flexible imitation learning systems capable of handling continuous action spaces.

\section{Related Work}

In this work, we focus on learning policies from demonstrations using primarily state observations rather than full state-action trajectories. This area is commonly known as \emph{imitation from observation} or \emph{state-only imitation learning}. It has seen significant growth in recent years, and we refer interested readers to recent surveys for comprehensive coverage (e.g.,~\cite{torabi2019,zare2024survey}). In brief, research in this area spans various approaches, including learning goals~\cite{liu2018imitation,sharma2019third,smith2019avid}, reward functions~\cite{xiong2021learning}, or trajectory waypoints~\cite{Wang2023MimicPlayLI} from state-only demonstrations. To enhance generalization across different visual settings and tasks, prior studies have focused on learning visual dynamics~\cite{yu2018one,Schmeckpeper2019LearningPM,schmeckpeper2020reinforcement} or visual representations~\cite{nair2022rm,lyu2023task}. Some approaches also incorporate auxiliary signals, such as rewards~\cite{zheng2023semi,schmeckpeper2020reinforcement} or interactions with the environment~\cite{torabi2018behavioral,radosavovic2021state,seo2022reinforcement}.

Our work is related to methods that learn inverse dynamics models. For example, VPT~\cite{baker2022video} trains an inverse dynamics model as an action labeler using labeled trajectories. VPT uses this model to assign pseudo-actions to action-free trajectories (videos), achieving human-level performance for Minecraft agents on specific tasks. However, unlike \method{}, VPT does not use a forward dynamics model and risks overfitting the action labeler when few action trajectories are available.

The most relevant works to ours are latent action learning methods ILPO~\cite{edwards2019imitating} and LAPO~\cite{schmidt2024learning}. Unlike \method{}, ILPO measures differences in the observation space without linear latent states and focuses on \emph{discrete actions}; its training loss requires marginalization over latent actions. Similar to \method{}, LAPO jointly models forward and inverse dynamics to learn latent actions, but it regularizes latent action learning using a vector-quantized information bottleneck~\cite{10.5555/3295222.3295378}. As vector quantization has difficulty scaling to high-dimensional continuous actions~\cite{mentzer2024finite}, \method{} takes a different approach by lifting the nonlinear system into a linear space, where the latent actions can be induced to directly capture state differences.

\section{Preliminaries}

\subsection{Problem Formulation}

We aim to learn a policy $\pi$ given a set of action-free expert demonstrations: $\mathcal{D}_x = \{(x^i_0,x^i_1,\cdots,x^i_T)\}_{i=1}^N$, alongside a smaller set of demonstrations with action labels: $\mathcal{D}_a = \{(x^i_0,a^i_0,x^i_1,a^i_1,\cdots,a^i_{T-1},x^i_T)\}_{i=1}^M$, where $M\ll N$. We wish to exploit the large action-free dataset to facilitate learning when only a small number of action-labeled trajectories are available. 

\subsection{Plan-then-Control with Diffusion Planners and Inverse Dynamics Controllers}

We adopt a \emph{plan-then-control} scheme, where a \emph{planner} generates $k$ future states conditioned on the current state $x_t$ and history $h_t$ (e.g., previous states): $\pi_{PL}(x_{t+1:t+k}|x_t,h_t)$.
The \emph{inverse dynamics controller} predicts actions that would give rise to these generated future states: $\pi_{C}(a_{t:t+k-1}|x_{t:t+k},h_{t})$. The combined system forms a policy $\pi(a_{t:t+k-1}|x_t,h_t)$ that generates action sequences conditioned on current state and history.

Our work makes use of diffusion model planners, which have been shown to be effective for planning~\cite{ajay2023is,liang2024dreamitaterealworldvisuomotorpolicy}. Early works have used diffusion models to directly predict actions ~\cite{janner2022diffuser,Chi_RSS_23,ajay2023is,prasad2024consistency,liang2024dreamitaterealworldvisuomotorpolicy,Chen-RSS-24}, but here, we diffuse intended future states, similar to ~\cite{ajay2023is}. This enables us to train the planner using only the action-free dataset $\mathcal{D}_x$. 
The remaining challenge and main problem addressed in this work is to learn an effective controller $\pi_C$ given limited action labels.

\subsection{Koopman Operator Theory}
In our work, we learn representations with linear dynamics, which finds connections to Koopman Operator Theory~\cite{koopman1931hamiltonian,koopman1932dynamical}. The key notion is to represent a non-linear dynamical system as a linear system by projecting (\emph{lifting}) the states into an infinite-dimensional space. 

More formally, let $x_t \in \mathcal{X} \subset \mathbb{R}^n$ denotes the state of a dynamical system at time $t$.
Given a discrete-time dynamical system governed by $x_{t+1} = \mathbf{F}(x_t)$, let $\mathcal{F}$ denote the collection of observable functions $g: \mathcal{X} \rightarrow \mathbb{R}$ that forms an infinite-dimensional Hilbert space. The Koopman Operator $\mathcal{K}:\mathcal{F}\rightarrow\mathcal{F}$ is defined as a linear transformation acting on these observables, such that $\mathcal{K}g \triangleq g\circ \mathbf{F}$ for any $g \in \mathcal{F}$. The Koopman Operator transforms observable functions according to the system dynamics, such that:
\begin{equation}
    \mathcal{K}g(x_{t}) =g(\mathbf{F}(x_t)) = g(x_{t+1}). 
\end{equation} 
Koopman Operators have been extended to nonlinear \emph{control} systems 

$x_{t+1} = \mathbf{F}(x_t,a_t)$ by linearly injecting the control input $a_t$ into the Koopman Observation space~ \cite{Brunton2015KoopmanIS,Bruder_Gillespie_David_Remy_Vasudevan_2019,brunton2016koopman},
\begin{equation}
    g(x_{t+1}) = \mathcal{K}g(x_t) + \mathcal{C}a_t
    \label{kpm_ctrl}
\end{equation}
where $\mathcal{C}$ is referred as the \textit{Control Matrix}. 

In practice, by identifying a finite-dimensional subspace where $\mathcal{K}$ remains invariant, the Koopman Operator can be represented as a \textit{Koopman Matrix} with finite dimension, $\mathcal{K}\in \mathbb{R}^{m\times m}$, where $g: \mathcal{X}\rightarrow \mathbb{R}^m$ can be hand-specified or learned~\cite{Lusch2017DeepLF,Shi2022DeepKO}. Deep Koopman Operators based on neural networks have been popular and applied successfully to various problems in robotics and control, including multi-object systems and soft robots~\cite{li2020learning}, long-horizon planning \cite{mondal2024efficient}, visual-motor policy learning \cite{lyu2023taskoriented}, dexterous hand manipulation \cite{han2023on}, optimal control \cite{Mamakoukas2019LocalKO}, and inverse optimal control \cite{Liang2023ADA}. Existing  learning methods typically assume the availability of abundant state-action data to learn the Koopman Operators and control matrices \emph{jointly}. Next, we will extend this framework to scenarios where action data is limited. 

\section{Imitation Learning with Limited Action Data}

In this work, we present an efficient controller assuming abundant state observations and limited action data. Our approach is to learn useful \emph{latent actions} that can be quickly mapped to \emph{continuous} real robot actions.

\method{} embodies a simple strategy: regularizing latent action learning with linear latent dynamics and linear action decoding. Enforcing linear forward dynamics in a lifted latent state space helps the learned latent actions capture state differences. This is in contrast to prior works like ILPO and LAPO, which model more complex relationships between latent actions and dynamics. 
Further regularizing the action decoder (e.g., to be linear) minimizes the risk of overfitting, particularly with limited action-labeled data. 

\para{Lifting the System and Latent Action Prediction.} We use Deep Koopman embeddings to lift the system and assume that sequences of consecutive latent states encode action/control information. Let $x^{\tau}_t$ represent a trajectory around time $t$, consisting of past, current, and a near future state $x_{t+k}$: $x^{\tau}_t = (x_{t-n}, x_{t-n+1}, \cdots, x_{t}, x_{t+k})$. 
The state trajectory encoder (latent action predictor) $f_\theta$ maps such trajectories to control signals in the Koopman observation space:
\begin{align}
    g_\theta(x_{t+1}) &= \mathcal{K}_\theta \times g_\theta(x_t) + f_\theta(x^{\tau}_t) \label{latent_act_kpm}
\end{align}

The design of state trajectory allows near future state to provide information about the action to take, but disallows direct feeding in the future state $x_{t+1}$, which prevents learning trivial representation that predicts $x_{t+1}$ directly.
The linear structure encourages $f_\theta$ to learn the difference between lifted states, $g_\theta(x_{t+1})$ and $\mathcal{K}_\theta \times g_\theta(x_t)$, without needing complex information bottlenecks. 
We assume state-action trajectories are generated by the same system, so the observation function $g_\theta$ remains the same.
We define an action decoder $d_\phi$:
\begin{equation}
    a_t = d_\phi(f_\theta(x^{\tau}_t)) \label{act_decoder}
\end{equation}
A linear decoder helps avoid overfitting when few action-labeled trajectories are available. While a natural question arises regarding the sufficiency of a linear model to remap latent actions to real robot actions, our experiments demonstrates linear decoding to be effective. However, more complex decoders can be introduced if performance declines (with the corresponding risk of overfitting).

\para{Training.} We jointly train the Koopman observation function $g_{\theta}$, Koopman matrix $\mathcal{K}_{\theta}$, and latent action predictor $f_{\theta}$. If action data is available, we also train the action decoder $d_{\phi}$. The observation function $g_{\theta}$ is trained as part of an autoencoder, with $g_{\theta}$ as the encoder and $g_{\psi}$ as the decoder:
\begin{equation}
    \mathcal{L}_{recon} = \frac{1}{T}\sum_{t=1}^T \| g_{\psi}(g_{\theta}(x_t)) - x_t\|^2
\end{equation}

We use a Koopman dynamics loss to learn $\mathcal{K}_{\theta}$ and trajectory encoder $f_{\theta}$,
\begin{equation}
    \mathcal{L}_{kpm} = \frac{1}{T}\sum_{t=1}^T \|\mathcal{K}_{\theta} \times g_{\theta}({x_t}) + f_{\theta}(x_t^{\tau}) - g_{\theta}({x_{t+1}})\|^2
\end{equation}

The action decoder $d_{\phi}$ is trained with an action prediction loss:
\begin{equation}
    \mathcal{L}_a = \frac{1}{T}\sum_{t=1}^T\| d_{\phi}(f_{\theta}(x_t^{\tau})) - a_t\|^2
\end{equation}

The total loss combines these terms: $\mathcal{L}_{total}= \mathcal{L}_{recon} + \lambda_1\mathcal{L}_{kpm}+ \lambda_2\mathcal{L}_a$, where $\lambda_1$ and $\lambda_2$ are hyperparameters. If no action labels are available, the action prediction loss $\mathcal{L}_a$ is omitted. We apply L2 regularization to all parameters.
\section{Experiments}
In this section, we report our experiment results using simulation benchmarks. We are interested in KOAP's effectiveness under varying amounts of action data and hypothesize that KOAP effectively learns latent actions, which results in good policy performance when action-labeled trajectories are limited.

\subsection{Benchmark Tasks}
We use the \textit{Datasets with Diverse human Demonstrations for Imitation Learning} (D3IL) benchmark~\cite{jia2024towards}, containing multi-modal human demonstrations and robot trajectories across five manipulation tasks of varying difficulty:
\begin{itemize}
    \item \textit{Avoiding}: Robot navigates from left to right while avoiding fixed obstacles. State includes end-effector position; actions control end-effector velocity. Collisions result in failure.
    
    \item \textit{Aligning}: Robot manipulates (push or pull) a block to match a target pose. State includes end-effector position, block pose, and target pose; actions control end-effector velocity.
    
    \item \textit{Sorting (4 blocks)}: Robot sorts four blocks (two red, two blue) into designated zones. State includes end-effector position and block poses; actions control end-effector velocity. Placing blocks in wrong zones causes failure.
    
    \item \textit{Stacking}: This is a difficult task. Robot stacks blocks in a specific order within a target zone. State includes end-effector position, gripper status, block poses, and stacking order; actions control 7-DoF joint velocity (which is not observed in state) and gripper.  Placing blocks outside the target zone or in the wrong order results in failure.

    \item \textit{Partially-Observable Sorting}: A more difficult partially-observable version of original Sorting task, where only block positions (not poses) are observable, with added Gaussian noise to end-effector positions in the state.

\end{itemize}
Due to big variation in dataset size across tasks, for training, we use 50\% of the action-labeled data for the  smaller task datasets (\textit{avoiding} and \textit{aligning}) and 25\% for larger datasets (\textit{sorting} and \textit{stacking}). We defined five different ``levels'' of action data based on dataset size. For the smaller tasks, the levels corresponded to (2\%, 5\%, 10\%, 25\%, 50\%), while for the larger tasks, the levels were (1\%, 2\%, 5\%, 10\%, 25\%).

\begin{figure}
    \centering
    \includegraphics[width=0.9\linewidth]{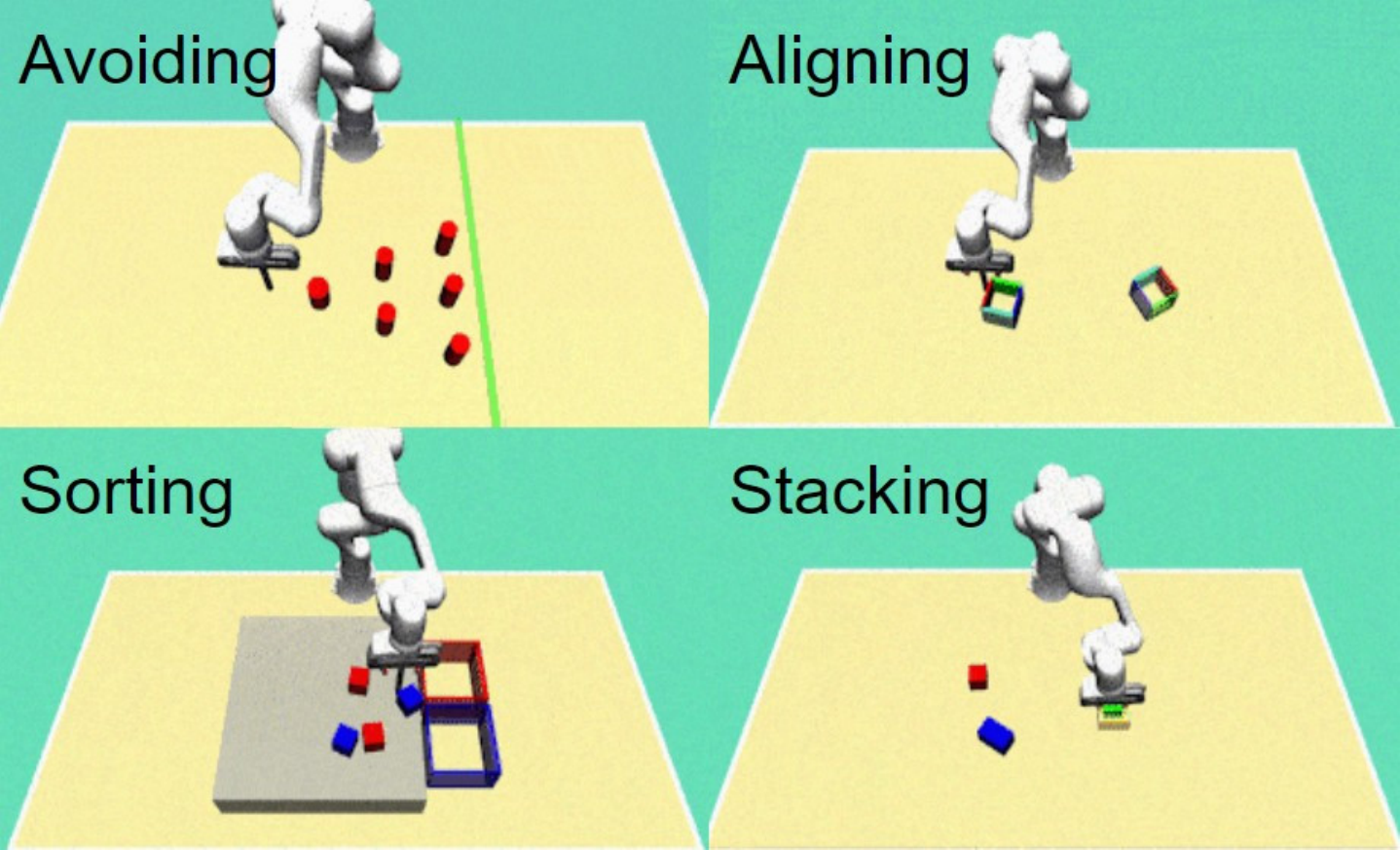}
    \caption{Original D3IL Tasks. The robot learns from a dataset containing partially-labeled multi-modal expert demonstrations. The goal is to manipulate object(s) to reach target positions or poses, adhering to task-specific rules (e.g., collision-free or color-based sorting).}
    \label{d3il}
\end{figure}

\subsection{Baselines}
In addition to \method, we ran several strong baseline methods on the above tasks:
\begin{itemize}
    \item \textbf{Diffusion Policy (DP)}~\cite{Chi_RSS_23}: A popular diffusion-based method for learning multi-modal behaviors, which uses action-labeled trajectories for training.
    
    \item \textbf{Decision Diffuser (DD)}~\cite{ajay2023is}: This method is similar to KOAP, but uses a pure supervised-learning approach for learning the controller. It allows us to better isolate the benefit of improving the inverse dynamics controller in a decision diffuser.

    \item \textbf{VAE}~\cite{Kingma2013AutoEncodingVB}: A Variational Autoencoder method that trains a controller by encoding full observational trajectories into an embedding. After training the VAE, an action decoder is trained to map sampled latent embeddings from the VAE to robot actions. This model can potentially learn useful action embeddings from the observations, but does not leverage dynamics.

    \item \textbf{LAPO}~\cite{schmidt2024learning}: We adopt the Decision Diffuser as above, but use LAPO to learn the inverse dynamics controller. Since no official implementation of LAPO is publicly available, we implemented LAPO following the paper with one key difference: the original LAPO paper uses VQ-VAE~\cite{van2017neural}, but we instead employ the more recent Finite Scalar Quantization (FSQ) method~\cite{mentzer2024finite}, which scales better to larger codebooks and is  more suitable for continuous action spaces; in preliminary trials,  FSQ outperformed VQ-VAE. All observational data is used for learning the dynamics models, while action-labeled data is used to learn the mapping from latent actions to real robot actions. 
    
\end{itemize}
All methods, except for Diffusion Policy, follow a plan-then-control scheme. All plan-then-control methods use the same Decision Diffuser for a given task. For fair comparison, we use the same underlying model for the Diffusion Policy and Decision Diffuser, with the difference being that the Diffusion Policy predicts future actions instead of states. For \method, $g_\theta$ and $g_\psi$ were MLPs and $f_\theta$ was a LSTM. The decoder $d_\phi$ was a linear model.  

We use the original dataset provided by D3IL and sample fixed-length trajectory sequences. The first two states (no actions) are used as conditioning input for the planner and controller. For future state $x_{t+k}$ in state trajectory, we find $k=8$ performs well empirically. For fair comparison, all controllers are trained to predict action chunks (with 12 steps). During inference, we employ receding-horizon control: the first 4 predicted actions are executed in the environment, with replanning and execution until the episode is completed. For complete details on the implementation, see our
\textcolor{blue}{\href{https://github.com/jxbi1010/KOAP/}{\text{online code repository}}}.

\begin{figure}
    \centering
    \includegraphics[scale=0.36]{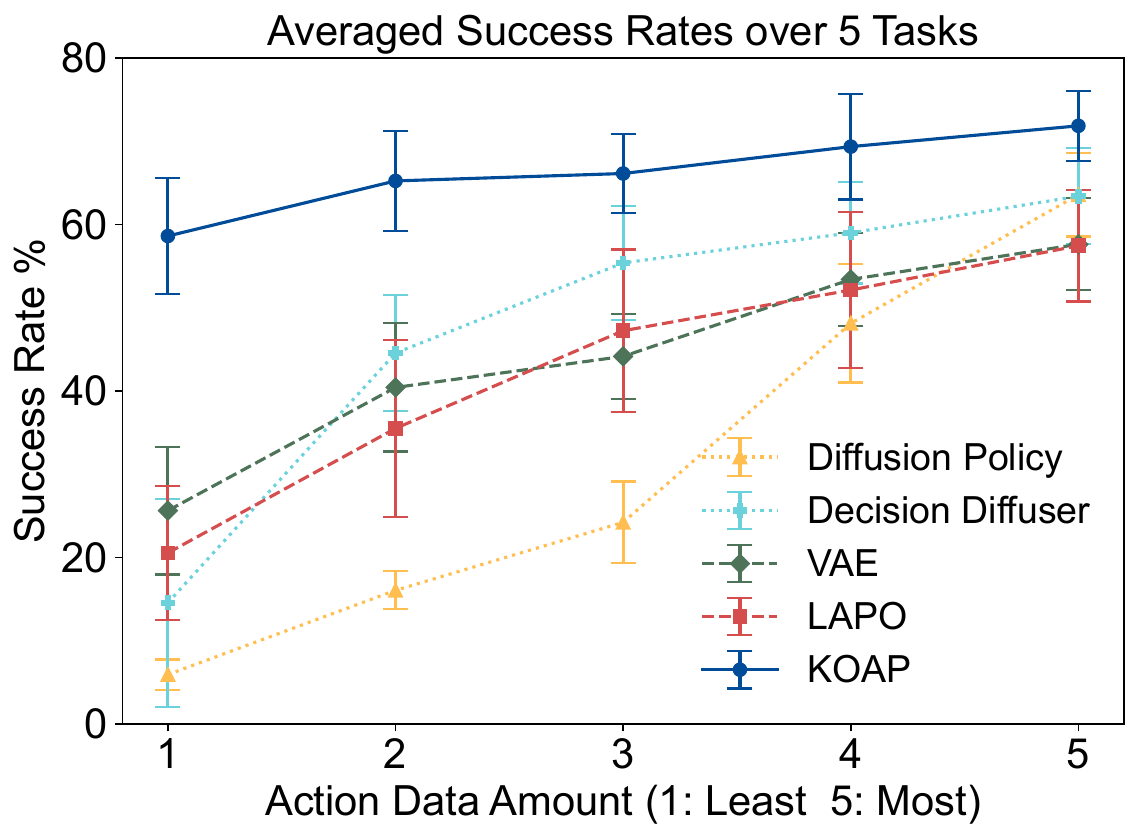}
    \caption{Average the success rates of each method versus the relative amount of action data used.}
    \label{fig:all_absolute}
\end{figure}

\begin{table*}[ht]
    \centering
    \caption{Task success rates for the different methods given varying amounts of action data (Levels 1 to 5 from smallest to largest). For Stacking, we report success rate of stacking 1 block at target position.}
    \begin{tabular}{c|c|cccccc}
        \toprule
        
        \textbf{Action Data Amt.}& \textbf{Task} &\textbf{DP} & \textbf{DD} & \textbf{VAE} & \textbf{LAPO} &\textbf{KOAP} (ours) \\
        \midrule
        \multirow{5}{*}{\textbf{1}} 
        &\textbf{Avoiding}  & $6.93\pm4.55$ &$36.70\pm11.80$  &$36.95\pm15.84$  &$43.63\pm12.57$  & $\mathbf{78.03\pm8.63}$\\
        &\textbf{Aligning}  &$10.55\pm1.08$ &$23.90\pm10.92$ &$38.00\pm4.33$ &$36.97\pm5.15$ &$\mathbf{80.28\pm4.94}$ \\
        &\textbf{Sorting}  &$0\pm0$          &$0\pm0$     &$36.80\pm8.03$ &$11.38\pm14.81$ &$\mathbf{61.65\pm5.63}$\\
        &\textbf{Stacking}  &$12.22\pm3.42$ &$12.22\pm9.70$ &$7.78\pm5.82$ &$6.67\pm5.36$ &$\mathbf{27.22\pm8.89}$\\
        &\textbf{PO-Sorting}  &$0\pm0$ &$0\pm0$ &$8.62\pm4.36$ &$1.11\pm2.48$ &$\mathbf{45.83\pm6.85}$\\
        \midrule
        
        \multirow{5}{*}{\textbf{2}} 
        &\textbf{Avoiding}  & $10.82\pm2.10$ &$77.78\pm7.11$  &$68.05\pm14.63$ &$70.82\pm12.66$  & $\mathbf{98.05\pm1.13}$ \\
        &\textbf{Aligning}  &$45.00\pm4.61$ &$72.23\pm7.93$ &$67.72\pm9.36$ &$60.85\pm8.75$ &$\mathbf{85.55\pm2.47}$ \\
        &\textbf{Sorting}  &$0.28\pm0.63$ &$49.67\pm7.57$ &$43.28\pm7.66$ &$25.55\pm16.79$ &$\mathbf{67.23\pm7.03}$\\
        &\textbf{Stacking}  &$24.18\pm4.00$ &$21.4\pm9.90$ &$9.17\pm3.68$ &$14.18\pm9.54$ &$\mathbf{26.38\pm13.53}$\\
        &\textbf{PO-Sorting}  &$0\pm0$ &$1.67\pm2.35$ &$13.88\pm3.14$ &$6.11\pm5.50$ &$\mathbf{48.89\pm5.67}$\\
        \midrule

        \multirow{5}{*}{\textbf{3}} 
        &\textbf{Avoiding}  & $21.12\pm6.44$ &$70.80\pm9.07$  &$64.18\pm4.27$  &$76.38\pm6.41$  & $\mathbf{97.22\pm2.49}$\\
        &\textbf{Aligning}  &$63.33\pm8.27$ &$83.62\pm4.00$ &$68.00\pm8.81$ &$75.57\pm7.37$ &$\mathbf{86.12\pm3.14}$ \\
        &\textbf{Sorting}  &$5.28\pm2.23$ &$\mathbf{64.38\pm5.05}$ &$55.80\pm3.30$ &$33.07\pm18.90$ &$61.38\pm6.07$\\
        &\textbf{Stacking}  &$28.60\pm5.31$  &$30.55\pm7.32$ &$16.38\pm7.23$ &$20.57\pm12.06$ &$\mathbf{33.88\pm7.73}$\\
        &\textbf{PO-Sorting}  &$2.78\pm2.26$ &$27.50\pm8.94$ &$16.38\pm1.93$ &$30.55\pm4.04$ &$\mathbf{51.95\pm4.35}$\\
        \midrule

        \multirow{5}{*}{\textbf{4}} 
        &\textbf{Avoiding}  & $78.90\pm7.68$ &$88.05\pm4.45$  &$77.77\pm9.68$ &$75.00\pm4.59$  & $\mathbf{98.33\pm1.92}$ \\
        &\textbf{Aligning}  &$77.77\pm7.98$ &$81.93\pm2.80$ &$83.80\pm4.11$ &$82.50\pm5.51$ &$\mathbf{88.35\pm3.19}$ \\
        &\textbf{Sorting} &$21.67\pm6.08$ &$62.72\pm6.57$ &$61.23\pm2.95$ &$44.18\pm15.49$ &$\mathbf{68.62\pm8.03}$\\
        &\textbf{Stacking} &$\mathbf{42.50\pm8.10}$  &$30.55\pm4.14$ &$23.33\pm5.36$ &$27.78\pm13.96$ &$36.95\pm10.80$\\
        &\textbf{PO-Sorting} &$19.72\pm5.64$ &$31.67\pm12.47$ &$20.82\pm5.59$ &$31.11\pm7.31$ &$\mathbf{54.44\pm7.80}$\\
        \midrule

        \multirow{5}{*}{\textbf{5}} 
        &\textbf{Avoiding}  & $94.70\pm1.78$  &$91.40\pm4.24$  &$86.37\pm3.52$  &$81.93\pm5.05$  & $\mathbf{97.50\pm1.58}$\\
        &\textbf{Aligning}  &$81.68\pm3.74$ &$81.08\pm7.05$ &$80.23\pm7.05$ &$82.77\pm4.70$ &$\mathbf{91.95\pm5.38}$ \\
        &\textbf{Sorting}  &$42.77\pm5.25$ &$66.08\pm4.46$ &$\mathbf{67.88\pm4.15}$ &$64.17\pm8.31$ &$65.83\pm3.68$\\
        &\textbf{Stacking} &$\mathbf{53.05\pm10.47}$ &$32.78\pm9.31$ &$30.83\pm9.98$ &$27.20\pm7.62$ &$45.57\pm5.43$\\
        &\textbf{PO-Sorting} &$45.58\pm3.82$ &$45.55\pm3.81$ &$22.76\pm2.85$ &$31.11\pm7.80$ &$\mathbf{58.33\pm5.00}$\\
        
        \bottomrule
    \end{tabular}
    \label{tab:tasks_transposed}
\end{table*}

\subsection{Results and Analysis}

\para{KOAP learns latent actions that are informative and results in strong policy performance.} 
Fig. \ref{fig:all_absolute} and Table \ref{tab:tasks_transposed} summarizes each method's policy success rate averaged across six different seeds, with 60 rollouts per seed (a total of 360 rollouts). \method{} significantly outperforms the standard Decision Diffuser, highlighting the value of learning latent actions from observations. Performance differences were most pronounced when action data was scarce ($<$ 5\%), with the gap narrowing as more action data became available, which is consistent with the expectation that more data allows for better inverse model learning. In \textit{PO-Sorting}, the baselines experience significant degradation relative to their performance in \textit{Sorting}, while \method{} retains good performance. This indicates that \method{} learns meaningful latent actions, even under partially-observability.

For these D3IL tasks, LAPO did not perform well, especially on the high-dimensional \emph{Stacking} task, potentially because information valuable for precise continuous action control was lost in the quantization bottleneck. In contrast, \method{} uses continuous representations throughout, with regularization imposed in the linear latent dynamics. Moreover, the VAE learner's relatively lower performance suggests that incorporating dynamics information is important for learning good latent actions.

\begin{figure}
    \centering
    \includegraphics[scale=0.070]{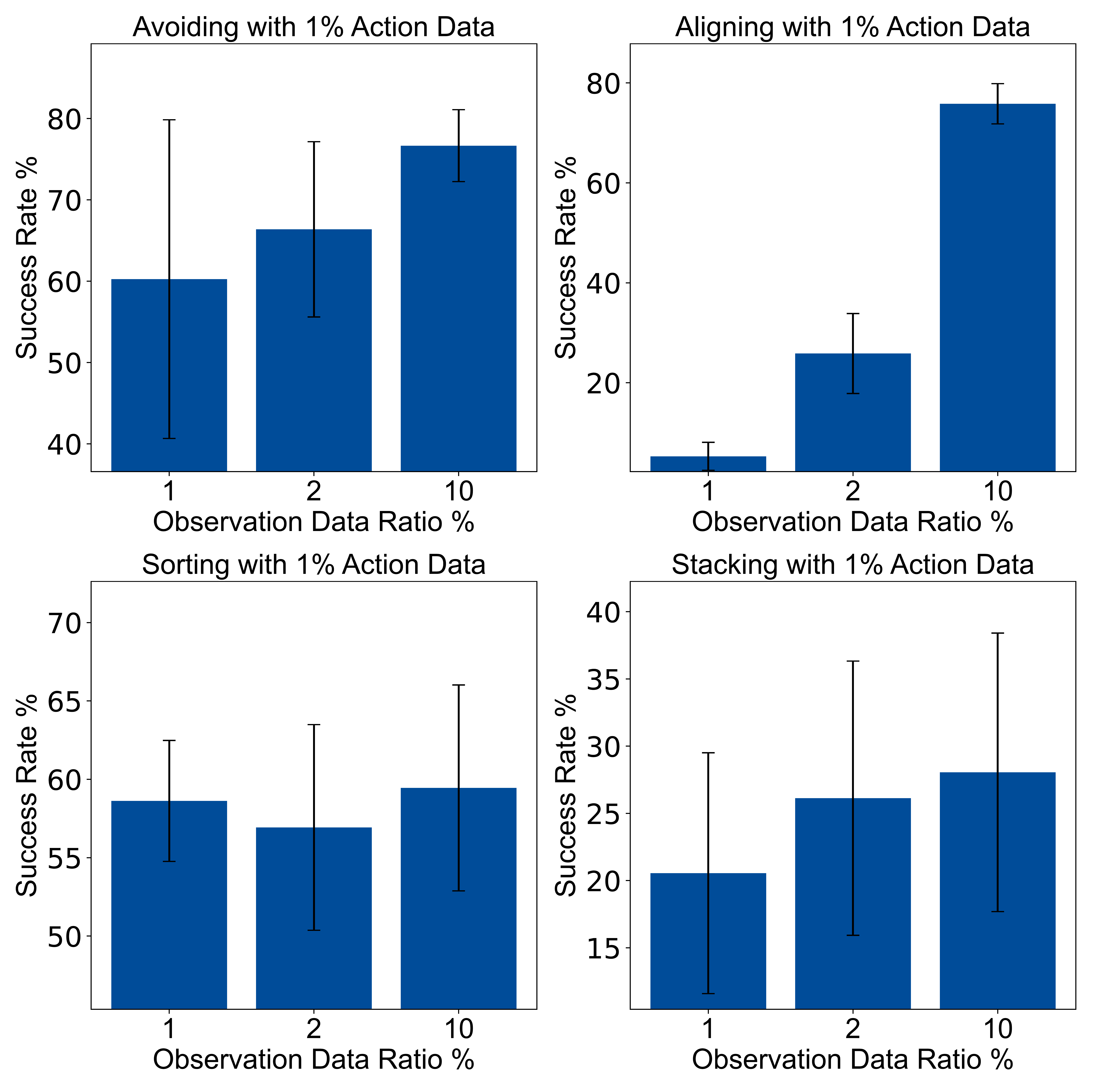}
    \caption{\method's performance increases with observation data.}
    \label{fig:kpm_obs}
    \vspace{-0.5cm}
\end{figure}

\para{KOAP learns better latent actions given more observational data.} Figure \ref{fig:kpm_obs} demonstrates a clear increase in success rates as more observation data is provided, while holding the action labels constant at 1\%. This result suggests that sufficient observation data is necessary for effective lifting of the nonlinear system and obtaining quality representations for latent action learning.

\para{Comparison with Variants.}
To better understand how \method's design influences its performance, we conducted experiments with several variants:
\begin{itemize}
    \item \textbf{Nonlinear}: Similar to KOAP but with nonlinear latent dynamics, where $g(x_{t+1}) = T_\theta(g(x_t), f(x^\tau))$, and $T_\theta$ is a neural network. It also uses a noninear action decoder.
    \item \textbf{Pretrain}: A two-stage approach where \method's latent actions are first trained without any action labels. Then, action labels are used to train the decoder and finetune \method's latent models (without access to the original observation data).
    \item \textbf{Relabel}: Inspired by ILPO~\cite{baker2022video}, this variant learns an action predictor \emph{without} a forward dynamics model and uses it to relabel the dataset, which is then used to train a standard Decision Diffuser.
\end{itemize}

\begin{table*}[ht]
    \centering
    \caption{Comparison of \method{} Variants showing Average Success Rates}
    \begin{tabular}{c|c|cccccc}
        \toprule
        \textbf{Action Data Amt.}&\textbf{Task}& \textbf{KOAP} & \textbf{Nonlinear} & \textbf{Pretrain} & \textbf{Relabel} \\
        \midrule
        \multirow{2}{*}{\textbf{1}} 
        &\textbf{Avoiding}  & $78.03\pm8.63$ &$64.17\pm13.97$  &$82.22\pm7.44$ &$5.83\pm10.96$ \\
        &\textbf{Stacking}  &$27.22\pm8.89$ &$14.17\pm8.86$ &$28.33\pm10.27$ & $4.17\pm6.07$\\
        \midrule
        \multirow{2}{*}{\textbf{2}} 
        &\textbf{Avoiding}  & $98.05\pm1.13$ &$70.00\pm10.80$  &$94.17\pm6.07$  &$57.50\pm11.09$\\
        &\textbf{Stacking}  &$26.38\pm13.53$ &$21.67\pm4.71$ &$27.50\pm7.50$ &$7.50\pm5.59$ \\
        \midrule
        \multirow{2}{*}{\textbf{3}} 
        &\textbf{Avoiding}  & $97.22\pm2.49$ &$83.33\pm8.50$ &$98.33\pm2.36$  &$65.83\pm16.18$\\
        &\textbf{Stacking}  &$33.88\pm7.73$ &$26.67\pm8.98$ &$32.50\pm9.47$ &$16.67\pm9.86$\\
        \midrule
        \multirow{2}{*}{\textbf{4}} 
        &\textbf{Avoiding}  & $98.33\pm1.92$ &$96.67\pm3.73$  &$98.33\pm2.36$  &$71.67\pm12.13$ \\
        &\textbf{Stacking}  &$36.95\pm10.80$ &$38.33\pm4.71$ &$38.33\pm10.27$ &$19.17\pm14.55$\\
        \midrule
        \multirow{2}{*}{\textbf{5}} 
        &\textbf{Avoiding}  &$97.50\pm1.58$  &$99.17\pm1.86$  &$93.33\pm12.80$  &$75.83\pm6.07$ \\
        &\textbf{Stacking}  &$45.57\pm5.43$ &$39.17\pm13.36$ &$47.50\pm9.46$ &$19.17\pm5.34$\\
        
        \bottomrule
    \end{tabular}
    \label{tab:ablations}
\end{table*}

Table \ref{tab:ablations} shows success rates (averaged over 6 seeds, each with 20 rollouts per method/task) for these variants on two D3IL tasks. We observe that \textbf{employing a nonlinear latent dynamics model was not as effective}, potentially because more parameters were introduced and the latent action learner was not longer forced to learn differences between latent states. The \textbf{relabeling variant also underperformed}, reinforcing the idea that observational data enhances latent action learning. Notably, even without any action labels, \method{} was able to \textbf{learn useful prior models that were easily finetuned} when action data becomes available.

\section{Case Study: Real Robot Scooping}
We evaluate \method{} on a robot scooping task, where the goal is to use a spoon to manipulate a chocolate piece in a bowl. The experiment was conducted using a 7-DoF Panda robot. 
Human demonstrations were collected by manually scooping the chocolate with the spoon, and robot action-labeled demonstrations were gathered through kinesthetic teaching using the same spoon. We collected 100 trajectories for both human and robot demonstrations. To reduce the computational overhead of generating pixel states for planning, we use a pretrained ResNet~\cite{7780459} to extract feature embeddings from pixel observations. The diffusion planner generates future embeddings conditioned on the current pixel observations, and the controller produces robot joint states as actions. 
\begin{figure}
    \centering
    \includegraphics[scale=0.50]{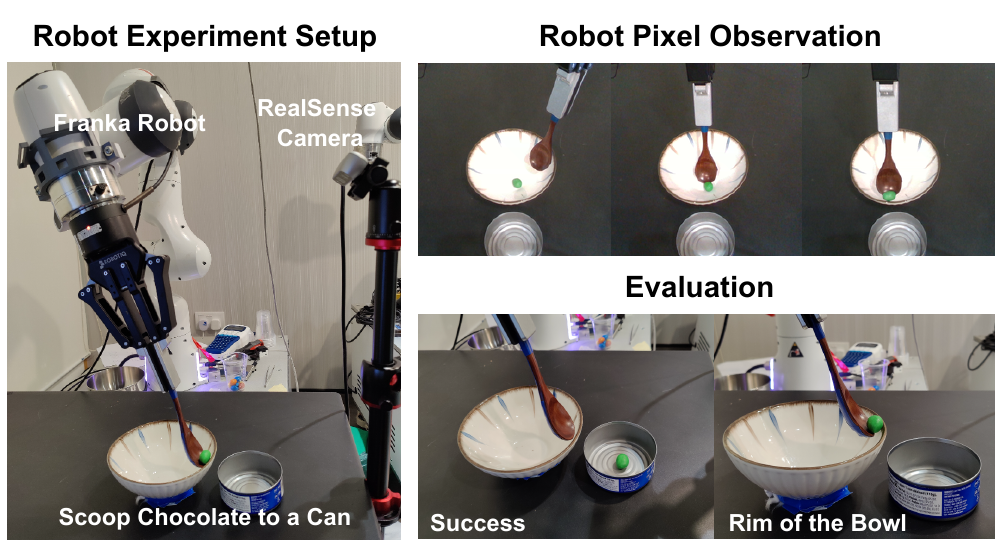}
    \caption{Real robot experiment setup: The robot starts by holding a spoon in a random initial position near the bowl. A camera provides a third-person view of the scene. Using pixel observations, the robot attempts to scoop the chocolate into the target container. We use two evaluation metrics: \emph{rim}, which indicates the robot was able to push the chocolate to the rim of the bowl, and \emph{success}, which indicates the chocolate was successfully scooped into the target container.}
    \label{fig:robot_setup}
    \vspace{-0.2cm}
\end{figure}
\begin{figure}
    \centering
    \includegraphics[scale=0.36]{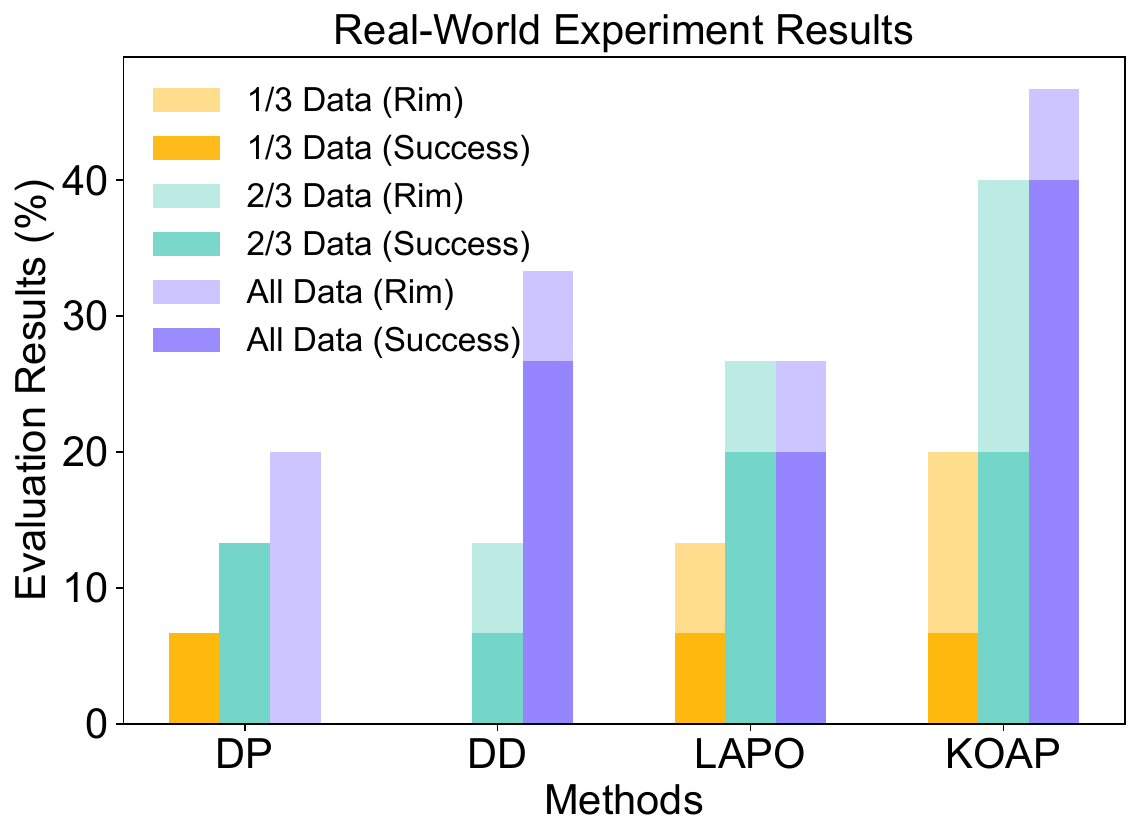}
    \caption{Performance across three different action data settings for the scooping task. We report success rates for both pushing the chocolate to the rim of the bowl and successfully scooping it into the target container.}
    \label{fig:robot_results}
    \vspace{-0.2cm}
\end{figure}

For each configuration, we ran 15 test episodes. Three random initial spoon positions were selected, and five episodes were conducted for each starting position. The success rates for the scooping task are shown in Fig. \ref{fig:robot_results}.

The results show that \method{} outperforms the baseline methods in both \emph{rim} and \emph{success} rates across varying amounts of action-labeled trajectory data. It significantly improves the standard Decision Diffuser and LAPO by learning a better controller. We observed the Diffusion Policy performed well at specific initial spoon configurations, but was not able to generalize to other locations. While there is room for improvement for all methods, these results shows that \method{} is able to leverage action-free demonstrations to enhance real robot performance.

\section{Conclusion}

In this work, we proposed the Deep Koopman Operator with Action Proxy (KOAP) method to enhance action-data efficiency in policy learning by leveraging observational data. Through simulated robot manipulation tasks, we demonstrated that KOAP effectively utilizes observational data to improve action-data efficiency in policy learning.
In addition, our real-world robot experiments showed that KOAP can successfully incorporate visual human demonstrations to facilitate visual-motor policy learning with minimal robot kinesthetic teaching trajectories. KOAP opens a new perspective for data-efficient policy learning, which could significantly contribute to the development of general-purpose robots in the future.

\section*{ACKNOWLEDGMENT}
This research is supported by the National Research Foundation, Singapore under its Medium Sized Center for Advanced Robotics Technology Innovation.

\bibliographystyle{IEEEtran}
\balance
\bibliography{reference}  

\begin{thebibliography}{10}
\providecommand{\url}[1]{#1}
\csname url@samestyle\endcsname
\providecommand{\newblock}{\relax}
\providecommand{\bibinfo}[2]{#2}
\providecommand{\BIBentrySTDinterwordspacing}{\spaceskip=0pt\relax}
\providecommand{\BIBentryALTinterwordstretchfactor}{4}
\providecommand{\BIBentryALTinterwordspacing}{\spaceskip=\fontdimen2\font plus
\BIBentryALTinterwordstretchfactor\fontdimen3\font minus
  \fontdimen4\font\relax}
\providecommand{\BIBforeignlanguage}[2]{{%
\expandafter\ifx\csname l@#1\endcsname\relax
\typeout{** WARNING: IEEEtran.bst: No hyphenation pattern has been}%
\typeout{** loaded for the language `#1'. Using the pattern for}%
\typeout{** the default language instead.}%
\else
\language=\csname l@#1\endcsname
\fi
#2}}
\providecommand{\BIBdecl}{\relax}
\BIBdecl

\bibitem{ajay2023is}
\BIBentryALTinterwordspacing
A.~Ajay, Y.~Du, A.~Gupta, J.~B. Tenenbaum, T.~S. Jaakkola, and P.~Agrawal, ``Is
  conditional generative modeling all you need for decision making?'' in
  \emph{The Eleventh International Conference on Learning Representations},
  2023. [Online]. Available: \url{https://openreview.net/forum?id=sP1fo2K9DFG}
\BIBentrySTDinterwordspacing

\bibitem{Lusch2017DeepLF}
\BIBentryALTinterwordspacing
B.~Lusch, J.~N. Kutz, and S.~L. Brunton, ``Deep learning for universal linear
  embeddings of nonlinear dynamics,'' \emph{Nature Communications}, vol.~9,
  2017. [Online]. Available:
  \url{https://api.semanticscholar.org/CorpusID:4854885}
\BIBentrySTDinterwordspacing

\bibitem{edwards2019imitating}
A.~Edwards, H.~Sahni, Y.~Schroecker, and C.~Isbell, ``Imitating latent policies
  from observation,'' in \emph{International conference on machine
  learning}.\hskip 1em plus 0.5em minus 0.4em\relax PMLR, 2019, pp. 1755--1763.

\bibitem{schmidt2024learning}
\BIBentryALTinterwordspacing
D.~Schmidt and M.~Jiang, ``Learning to act without actions,'' in \emph{The
  Twelfth International Conference on Learning Representations}, 2024.
  [Online]. Available: \url{https://openreview.net/forum?id=rvUq3cxpDF}
\BIBentrySTDinterwordspacing

\bibitem{jia2024towards}
\BIBentryALTinterwordspacing
X.~Jia, D.~Blessing, X.~Jiang, M.~Reuss, A.~Donat, R.~Lioutikov, and
  G.~Neumann, ``Towards diverse behaviors: A benchmark for imitation learning
  with human demonstrations,'' in \emph{The Twelfth International Conference on
  Learning Representations}, 2024. [Online]. Available:
  \url{https://openreview.net/forum?id=6pPYRXKPpw}
\BIBentrySTDinterwordspacing

\bibitem{torabi2019}
\BIBentryALTinterwordspacing
F.~Torabi, G.~Warnell, and P.~Stone, ``Recent advances in imitation learning
  from observation,'' in \emph{Proceedings of the Twenty-Eighth International
  Joint Conference on Artificial Intelligence, {IJCAI-19}}.\hskip 1em plus
  0.5em minus 0.4em\relax International Joint Conferences on Artificial
  Intelligence Organization, 7 2019, pp. 6325--6331. [Online]. Available:
  \url{https://doi.org/10.24963/ijcai.2019/882}
\BIBentrySTDinterwordspacing

\bibitem{zare2024survey}
M.~Zare, P.~M. Kebria, A.~Khosravi, and S.~Nahavandi, ``A survey of imitation
  learning: Algorithms, recent developments, and challenges,'' \emph{IEEE
  Transactions on Cybernetics}, 2024.

\bibitem{liu2018imitation}
Y.~Liu, A.~Gupta, P.~Abbeel, and S.~Levine, ``Imitation from observation:
  Learning to imitate behaviors from raw video via context translation,'' in
  \emph{2018 IEEE international conference on robotics and automation
  (ICRA)}.\hskip 1em plus 0.5em minus 0.4em\relax IEEE, 2018, pp. 1118--1125.

\bibitem{sharma2019third}
P.~Sharma, D.~Pathak, and A.~Gupta, ``Third-person visual imitation learning
  via decoupled hierarchical controller,'' \emph{Advances in Neural Information
  Processing Systems}, vol.~32, 2019.

\bibitem{smith2019avid}
L.~Smith, N.~Dhawan, M.~Zhang, P.~Abbeel, and S.~Levine, ``Avid: Learning
  multi-stage tasks via pixel-level translation of human videos,'' \emph{arXiv
  preprint arXiv:1912.04443}, 2019.

\bibitem{xiong2021learning}
H.~Xiong, Q.~Li, Y.-C. Chen, H.~Bharadhwaj, S.~Sinha, and A.~Garg, ``Learning
  by watching: Physical imitation of manipulation skills from human videos,''
  in \emph{2021 IEEE/RSJ International Conference on Intelligent Robots and
  Systems (IROS)}.\hskip 1em plus 0.5em minus 0.4em\relax IEEE, 2021, pp.
  7827--7834.

\bibitem{Wang2023MimicPlayLI}
\BIBentryALTinterwordspacing
C.~Wang, L.~J. Fan, J.~Sun, R.~Zhang, L.~Fei-Fei, D.~Xu, Y.~Zhu, and
  A.~Anandkumar, ``Mimicplay: Long-horizon imitation learning by watching human
  play,'' in \emph{Conference on Robot Learning}, 2023. [Online]. Available:
  \url{https://api.semanticscholar.org/CorpusID:257205825}
\BIBentrySTDinterwordspacing

\bibitem{yu2018one}
T.~Yu, C.~Finn, A.~Xie, S.~Dasari, T.~Zhang, P.~Abbeel, and S.~Levine,
  ``One-shot imitation from observing humans via domain-adaptive
  meta-learning,'' \emph{arXiv preprint arXiv:1802.01557}, 2018.

\bibitem{Schmeckpeper2019LearningPM}
\BIBentryALTinterwordspacing
K.~Schmeckpeper, A.~Xie, O.~Rybkin, S.~Tian, K.~Daniilidis, S.~Levine, and
  C.~Finn, ``Learning predictive models from observation and interaction,'' in
  \emph{European Conference on Computer Vision}, 2019. [Online]. Available:
  \url{https://api.semanticscholar.org/CorpusID:209515451}
\BIBentrySTDinterwordspacing

\bibitem{schmeckpeper2020reinforcement}
K.~Schmeckpeper, O.~Rybkin, K.~Daniilidis, S.~Levine, and C.~Finn,
  ``Reinforcement learning with videos: Combining offline observations with
  interaction,'' \emph{arXiv preprint arXiv:2011.06507}, 2020.

\bibitem{nair2022rm}
\BIBentryALTinterwordspacing
S.~Nair, A.~Rajeswaran, V.~Kumar, C.~Finn, and A.~Gupta, ``R3m: A universal
  visual representation for robot manipulation,'' in \emph{6th Annual
  Conference on Robot Learning}, 2022. [Online]. Available:
  \url{https://openreview.net/forum?id=tGbpgz6yOrI}
\BIBentrySTDinterwordspacing

\bibitem{lyu2023task}
X.~Lyu, H.~Hu, S.~Siriya, Y.~Pu, and M.~Chen, ``Task-oriented koopman-based
  control with contrastive encoder,'' in \emph{Conference on Robot
  Learning}.\hskip 1em plus 0.5em minus 0.4em\relax PMLR, 2023, pp. 93--105.

\bibitem{zheng2023semi}
Q.~Zheng, M.~Henaff, B.~Amos, and A.~Grover, ``Semi-supervised offline
  reinforcement learning with action-free trajectories,'' in
  \emph{International conference on machine learning}.\hskip 1em plus 0.5em
  minus 0.4em\relax PMLR, 2023, pp. 42\,339--42\,362.

\bibitem{torabi2018behavioral}
F.~Torabi, G.~Warnell, and P.~Stone, ``Behavioral cloning from observation,''
  in \emph{Proceedings of the 27th International Joint Conference on Artificial
  Intelligence}, 2018, pp. 4950--4957.

\bibitem{radosavovic2021state}
I.~Radosavovic, X.~Wang, L.~Pinto, and J.~Malik, ``State-only imitation
  learning for dexterous manipulation,'' in \emph{2021 IEEE/RSJ International
  Conference on Intelligent Robots and Systems (IROS)}.\hskip 1em plus 0.5em
  minus 0.4em\relax IEEE, 2021, pp. 7865--7871.

\bibitem{seo2022reinforcement}
Y.~Seo, K.~Lee, S.~L. James, and P.~Abbeel, ``Reinforcement learning with
  action-free pre-training from videos,'' in \emph{International Conference on
  Machine Learning}.\hskip 1em plus 0.5em minus 0.4em\relax PMLR, 2022, pp.
  19\,561--19\,579.

\bibitem{baker2022video}
B.~Baker, I.~Akkaya, P.~Zhokov, J.~Huizinga, J.~Tang, A.~Ecoffet, B.~Houghton,
  R.~Sampedro, and J.~Clune, ``Video pretraining (vpt): Learning to act by
  watching unlabeled online videos,'' \emph{Advances in Neural Information
  Processing Systems}, vol.~35, pp. 24\,639--24\,654, 2022.

\bibitem{10.5555/3295222.3295378}
A.~van~den Oord, O.~Vinyals, and K.~Kavukcuoglu, ``Neural discrete
  representation learning,'' in \emph{Proceedings of the 31st International
  Conference on Neural Information Processing Systems}, ser. NIPS'17.\hskip 1em
  plus 0.5em minus 0.4em\relax Red Hook, NY, USA: Curran Associates Inc., 2017,
  p. 6309–6318.

\bibitem{mentzer2024finite}
\BIBentryALTinterwordspacing
F.~Mentzer, D.~Minnen, E.~Agustsson, and M.~Tschannen, ``Finite scalar
  quantization: {VQ}-{VAE} made simple,'' in \emph{The Twelfth International
  Conference on Learning Representations}, 2024. [Online]. Available:
  \url{https://openreview.net/forum?id=8ishA3LxN8}
\BIBentrySTDinterwordspacing

\bibitem{liang2024dreamitaterealworldvisuomotorpolicy}
\BIBentryALTinterwordspacing
J.~Liang, R.~Liu, E.~Ozguroglu, S.~Sudhakar, A.~Dave, P.~Tokmakov, S.~Song, and
  C.~Vondrick, ``Dreamitate: Real-world visuomotor policy learning via video
  generation,'' 2024. [Online]. Available:
  \url{https://arxiv.org/abs/2406.16862}
\BIBentrySTDinterwordspacing

\bibitem{janner2022diffuser}
M.~Janner, Y.~Du, J.~Tenenbaum, and S.~Levine, ``Planning with diffusion for
  flexible behavior synthesis,'' in \emph{International Conference on Machine
  Learning}, 2022.

\bibitem{Chi_RSS_23}
C.~Chi, S.~Feng, Y.~Du, Z.~Xu, E.~Cousineau, B.~C. Burchfiel, and S.~Song,
  ``{Diffusion Policy: Visuomotor Policy Learning via Action Diffusion},'' in
  \emph{Proceedings of Robotics: Science and Systems}, Daegu, Republic of
  Korea, July 2023.

\bibitem{prasad2024consistency}
A.~Prasad, K.~Lin, J.~Wu, L.~Zhou, and J.~Bohg, ``Consistency policy:
  Accelerated visuomotor policies via consistency distillation,'' in
  \emph{Robotics: Science and Systems}, 2024.

\bibitem{Chen-RSS-24}
K.~Chen, E.~Lim, L.~Kelvin, Y.~Chen, and H.~Soh, ``{Don't Start From Scratch:
  Behavioral Refinement via Interpolant-based Policy Diffusion},'' in
  \emph{Proceedings of Robotics: Science and Systems}, Delft, Netherlands, July
  2024.

\bibitem{koopman1931hamiltonian}
B.~O. Koopman, ``Hamiltonian systems and transformation in hilbert space,''
  \emph{Proceedings of the National Academy of Sciences}, vol.~17, no.~5, pp.
  315--318, 1931.

\bibitem{koopman1932dynamical}
B.~O. Koopman and J.~v. Neumann, ``Dynamical systems of continuous spectra,''
  \emph{Proceedings of the National Academy of Sciences}, vol.~18, no.~3, pp.
  255--263, 1932.

\bibitem{Brunton2015KoopmanIS}
\BIBentryALTinterwordspacing
S.~L. Brunton, B.~W. Brunton, J.~L. Proctor, and J.~N. Kutz, ``Koopman
  invariant subspaces and finite linear representations of nonlinear dynamical
  systems for control,'' \emph{PLoS ONE}, vol.~11, 2015. [Online]. Available:
  \url{https://api.semanticscholar.org/CorpusID:7675653}
\BIBentrySTDinterwordspacing

\bibitem{Bruder_Gillespie_David_Remy_Vasudevan_2019}
\BIBentryALTinterwordspacing
D.~Bruder, B.~Gillespie, C.~David~Remy, and R.~Vasudevan, ``Modeling and
  control of soft robots using the koopman operator and model predictive
  control,'' \emph{Robotics: Science and Systems XV}, Jun. 2019. [Online].
  Available: \url{http://www.roboticsproceedings.org/rss15/p60.pdf}
\BIBentrySTDinterwordspacing

\bibitem{brunton2016koopman}
S.~L. Brunton, B.~W. Brunton, J.~L. Proctor, and J.~N. Kutz, ``Koopman
  invariant subspaces and finite linear representations of nonlinear dynamical
  systems for control,'' \emph{PloS one}, vol.~11, no.~2, p. e0150171, 2016.

\bibitem{Shi2022DeepKO}
\BIBentryALTinterwordspacing
H.~bin Shi and M.~Q. Meng, ``Deep koopman operator with control for nonlinear
  systems,'' \emph{IEEE Robotics and Automation Letters}, vol.~7, pp.
  7700--7707, 2022. [Online]. Available:
  \url{https://api.semanticscholar.org/CorpusID:246867015}
\BIBentrySTDinterwordspacing

\bibitem{li2020learning}
\BIBentryALTinterwordspacing
Y.~Li, H.~He, J.~Wu, D.~Katabi, and A.~Torralba, ``Learning compositional
  koopman operators for model-based control,'' in \emph{International
  Conference on Learning Representations}, 2020. [Online]. Available:
  \url{https://openreview.net/forum?id=H1ldzA4tPr}
\BIBentrySTDinterwordspacing

\bibitem{mondal2024efficient}
\BIBentryALTinterwordspacing
A.~K. Mondal, S.~S. Panigrahi, S.~Rajeswar, K.~Siddiqi, and S.~Ravanbakhsh,
  ``Efficient dynamics modeling in interactive environments with koopman
  theory,'' in \emph{The Twelfth International Conference on Learning
  Representations}, 2024. [Online]. Available:
  \url{https://openreview.net/forum?id=fkrYDQaHOJ}
\BIBentrySTDinterwordspacing

\bibitem{lyu2023taskoriented}
\BIBentryALTinterwordspacing
X.~Lyu, H.~Hu, S.~Siriya, Y.~Pu, and M.~Chen, ``Task-oriented koopman-based
  control with contrastive encoder,'' in \emph{7th Annual Conference on Robot
  Learning}, 2023. [Online]. Available:
  \url{https://openreview.net/forum?id=q0VAoefCI2}
\BIBentrySTDinterwordspacing

\bibitem{han2023on}
\BIBentryALTinterwordspacing
Y.~Han, M.~Xie, Y.~Zhao, and H.~Ravichandar, ``On the utility of koopman
  operator theory in learning dexterous manipulation skills,'' in \emph{7th
  Annual Conference on Robot Learning}, 2023. [Online]. Available:
  \url{https://openreview.net/forum?id=pw-OTIYrGa}
\BIBentrySTDinterwordspacing

\bibitem{Mamakoukas2019LocalKO}
\BIBentryALTinterwordspacing
G.~Mamakoukas, M.~L. Casta{\~n}o, X.~Tan, and T.~D. Murphey, ``Local koopman
  operators for data-driven control of robotic systems,'' \emph{Robotics:
  Science and Systems XV}, 2019. [Online]. Available:
  \url{https://api.semanticscholar.org/CorpusID:197640648}
\BIBentrySTDinterwordspacing

\bibitem{Liang2023ADA}
\BIBentryALTinterwordspacing
Z.~Liang, W.~Hao, and S.~Mou, ``A data-driven approach for inverse optimal
  control,'' \emph{2023 62nd IEEE Conference on Decision and Control (CDC)},
  pp. 3632--3637, 2023. [Online]. Available:
  \url{https://api.semanticscholar.org/CorpusID:257913681}
\BIBentrySTDinterwordspacing

\bibitem{Kingma2013AutoEncodingVB}
\BIBentryALTinterwordspacing
D.~P. Kingma and M.~Welling, ``Auto-encoding variational bayes,'' \emph{CoRR},
  vol. abs/1312.6114, 2013. [Online]. Available:
  \url{https://api.semanticscholar.org/CorpusID:216078090}
\BIBentrySTDinterwordspacing

\bibitem{van2017neural}
A.~Van Den~Oord, O.~Vinyals \emph{et~al.}, ``Neural discrete representation
  learning,'' \emph{Advances in neural information processing systems},
  vol.~30, 2017.

\bibitem{7780459}
K.~He, X.~Zhang, S.~Ren, and J.~Sun, ``Deep residual learning for image
  recognition,'' in \emph{2016 IEEE Conference on Computer Vision and Pattern
  Recognition (CVPR)}, 2016, pp. 770--778.

\end{thebibliography}

\end{document}